\documentclass[10pt,twocolumn,letterpaper]{article}

\usepackage[pagenumbers]{cvpr} 

\newcommand{\red}[1]{{\color{red}#1}}

\usepackage{algorithm}
\usepackage{algpseudocode}
\usepackage{booktabs}
\usepackage{multirow}
\usepackage{tabularx,array}
\usepackage{graphicx}
\usepackage{xcolor}     
\newcolumntype{C}{>{\centering\arraybackslash}X} 

\definecolor{cvprblue}{rgb}{0.21,0.49,0.74}
\usepackage[pagebackref,breaklinks,colorlinks,allcolors=cvprblue]{hyperref}

\def\paperID{41623} 
\def\confName{CVPR}
\def\confYear{2026}

\title{DRiffusion: Draft-and-Refine Process Parallelizes Diffusion Models with Ease}

\author{Runsheng Bai\thanks{Work done while the author was at Tsinghua University.}\\
\textit{CSAIL, MIT}\\
{\tt\small runsheng@mit.edu}
\and
Chengyu Zhang\\
\textit{IST, Nanjing University}\\
{\tt\small zhangchengyu@smail.nju.edu.cn}
\and
Yangdong Deng\\
\textit{School of Software, Tsinghua University}\\
\textit{FuturististAI Lab, Shanghai Tsinghua International Innovation Center}\\
{\tt\small dengyd@tsinghua.edu.cn}
}

\begin{document}
\maketitle
\begin{abstract}
Diffusion models have achieved remarkable success in generating high-fidelity content but suffer from slow, iterative sampling, resulting in high latency that limits their use in interactive applications. We introduce DRiffusion, a parallel sampling framework that parallelizes diffusion inference through a draft-and-refine process. DRiffusion employs skip transitions to generate multiple draft states for future timesteps and computes their corresponding noises in parallel, which are then used in the standard denoising process to produce refined results. Theoretically, our method achieves an acceleration rate of $\tfrac{1}{n}$ or $\tfrac{2}{n+1}$, depending on whether the conservative or aggressive mode is used, where $n$ denotes the number of devices. Empirically, DRiffusion attains 1.4×–3.7× speedup across multiple diffusion models while incur minimal degradation in generation quality: on MS-COCO dataset, both FID and CLIP remain largely on par with those of the original model, while PickScore and HPSv2.1 show only minor average drops of 0.17 and 0.43, respectively. These results verify that DRiffusion delivers substantial acceleration and preserves perceptual quality.
\end{abstract}    
\section{Introduction}
\label{sec:intro}

The rise of diffusion models \cite{ho2020denoising, song2020score, kingma2021variational} has marked a significant evolution in generative modeling in recent years. These models have established themselves as a foundational technique and showcased their adaptability in applications ranging from image \& video generation \cite{rombach2022high, saharia2022photorealistic, betker2023improving, ho2022video, brooks2024video}, audio synthesis \cite{kong2020diffwave, huang2023make, liu2023audioldm} to molecular discovery \cite{xu2022geodiff, corso2022diffdock, hua2024mudiff}. The capability to produce such high-quality results stems from the core idea behind diffusion models: a repeated, iterative denoising process. This process typically begins with pure noise and progressively removes a fraction of the noise to reveal the underlying data structure more clearly. This iterative refinement ensures remarkable output fidelity but comes at the cost of significantly slower sampling speeds, hindering their broader applicability.

\begin{figure}[t]
  \centering
  \setlength{\tabcolsep}{3pt}

  \begin{tabular}{@{} m{.05\linewidth} @ {} m{.29\linewidth} m{.29\linewidth} m{.29\linewidth} @{} m{.05\linewidth} @{}}

    & \centering\includegraphics[width=\linewidth]{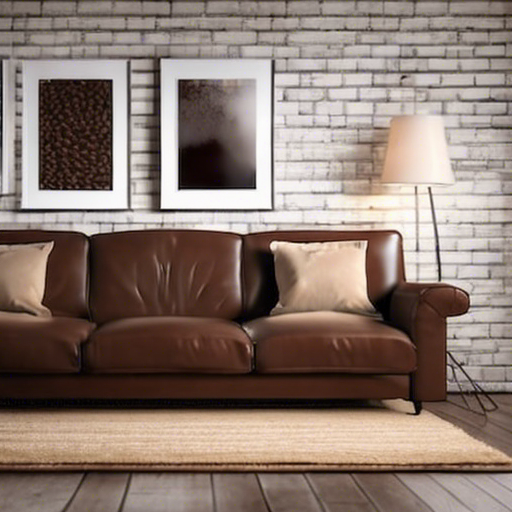} &
    \centering\includegraphics[width=\linewidth]{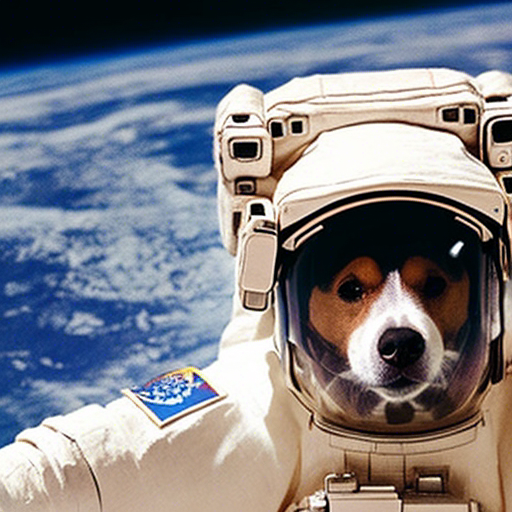} &
    \centering\includegraphics[width=\linewidth]{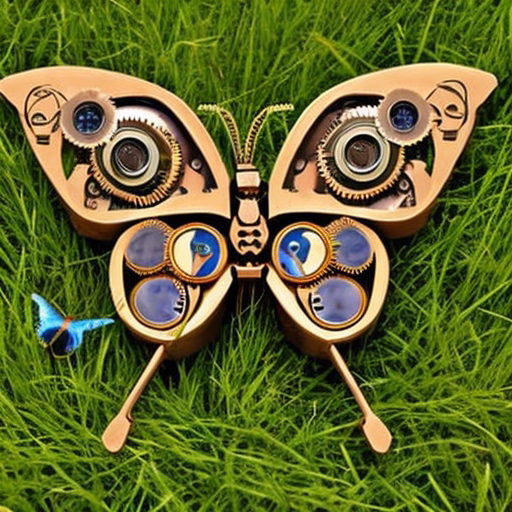} &
    \multirow{3}{*}{\centering\rotatebox[origin=c]{270}{\footnotesize\textbf{\quad \quad \red{2.8$\times$} Speed up}}}
    \\
    & \centering\includegraphics[width=\linewidth]{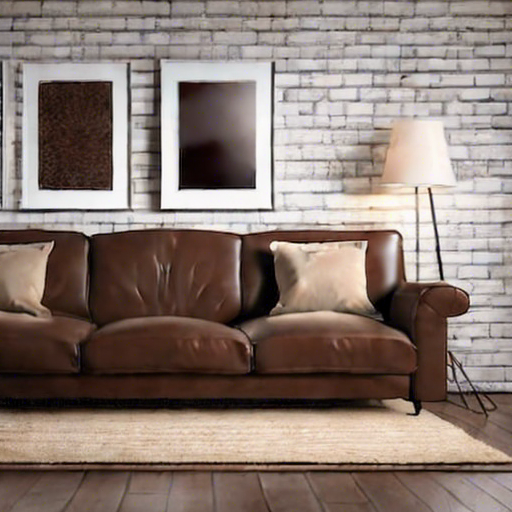} &
    \centering\includegraphics[width=\linewidth]{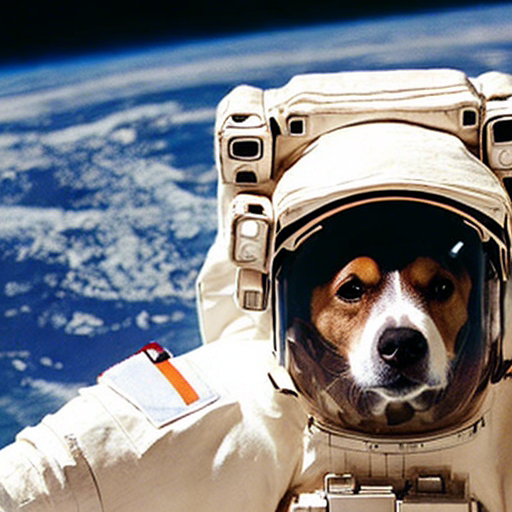} &
    \centering\includegraphics[width=\linewidth]{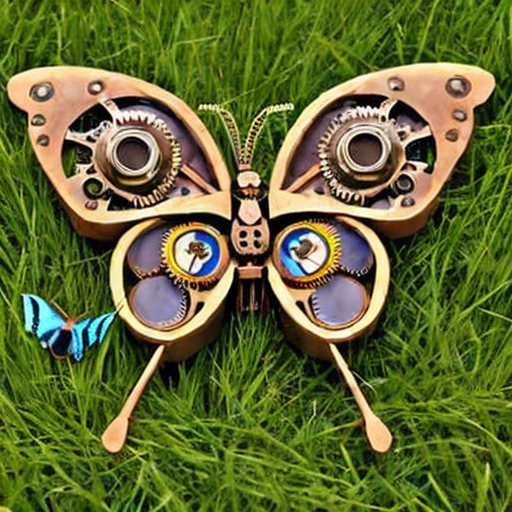} &
    \\
    & \centering\scriptsize \textit{A cozy living room with a brown leather sofa, a wooden coffee table, photo-realistic style.} &
    \centering\scriptsize \textit{A dog wearing an astronaut suit floating in outer space, Earth visible behind.} &
    \centering\scriptsize \textit{A steampunk-style mechanical butterfly resting on grass field.} &
    \\

  \end{tabular}

  \caption{Comparison across different prompts. Top row: original outputs; bottom row: results of 3-device aggressive parallelization.}
  \label{fig:comp}
\end{figure}

Significant efforts have been devoted to addressing this challenge. Approaches such as Rectified Flow \cite{lipman2022flow, liu2022flow, zhu2024slimflow} and distillation \cite{salimans2022progressive, luo2023latent} have achieved notable acceleration. Yet, when the number of sampling steps is reduced aggressively, these methods often incur noticeable trade-offs in output quality; and distillation, in particular, can substantially reduce generation diversity. Parallelization offers a promising alternative, as well as an orthogonal complement to these methods, yet existing implementations face significant limitations. System-level methods \cite{li2024distrifusion, chen2024asyncdiff}, designed from a computational perspective, tend to be constrained by the structural dependencies of the underlying network, i.e., specific to the models' architectures such as U-Net\cite{ronneberger2015u} or Transformers\cite{vaswani2017attention}. Mathematical approaches\cite{shih2023parallel, tang2024accelerating, lu2025parasolver}, which frame diffusion as Stochastic Differential Equation (SDE) or Ordinary Differential Equation (ODE) formulations and derive alternative solvers, often result in poor compatibility with current frameworks and may deviate from the original model’s sampling distribution significantly.

Instead, our investigation begins with a simple yet fundamental question: why is parallelizing diffusion models so challenging? The answer lies in the inherently sequential nature of the diffusion sampling process. Consequently, system-level methods seek computational elements that can be parallelized outside this sequence, while mathematical approaches attempt to redefine the entire sampling path. This leads us to a critical question: does the original framework contain any inherent parallelism at all? We demonstrate that, by leveraging a simple mathematical observation and a creative scheduling pattern, parallelism can be unlocked. Our answer, fortunately, is a definitive yes.

In this paper, we present \textbf{DRiffusion} (Draft-and-Refine Diffusion), an innovative method for parallelizing diffusion models that integrates advantages from both system-level and mathematical approaches. Our method starts with skip transitions within the sampling chain, and the core is to leverage these skips to draft approximate future states (e.g., $x_{t-2}, x_{t-3}$), which are then fed into the original pre-trained noise prediction network in parallel to obtain the predicted noises (or initial states). These noises are accurate, as the skip transitions are theoretically valid and the network exhibits strong generalization. Therefore, with these noises we can replay the original denoising updates from the current anchor to get the refined and accurate final outputs. As a result, we successfully unlock the underlying parallelism by exploiting skip transitions, thereby consolidating the primary computational bottleneck—the network prediction—into a single parallel step.

Figure \ref{fig:comp} shows that our DRiffusion method generates outputs consistent with original sampler and preserves the fine-grained details: for example, wrinkles on the surface of the leather sofa and natural shading on the dog's face. The inference speed is also significantly accelerated, as the method achieves nearly 3$\times$ wall-clock speedup with 3 GPUs. In summary, DRiffusion couples a straightforward theoretical property with a practical parallel implementation, delivering high-quality empirical results.

\section{Related Works}
\paragraph{Diffusion Models} Diffusion probabilistic models were first introduced by Sohl-Dickstein et al. \cite{sohl2015deep}, who formalized the forward–reverse noising framework. Building on this, Ho et al. \cite{ho2020denoising} introduce a shared noise-prediction denoiser and a discrete reverse-diffusion sampler in replace of the original variational setup, which then became the standard recipe for diffusion models. Subsequently, DDIM \cite{song2020denoising} was introduced by Song et al., expanding DDPM’s single stochastic Markov chain into a family of non-Markovian samplers that preserve the same marginals and improve sampling efficiency. In parallel, Song et al. \cite{song2020score} generalized the framework to continuous-time score-based SDEs and their probability-flow ODE counterparts, providing a unifying lens and tools for analysis.

\paragraph{Parallelism} As previously mentioned, parallelization of diffusion models can be roughly categorized into two different paths: systematical and mathematical methods. Systematical methods often spot the parallelizable component of the whole computation process and exploit it for efficiency. For example, DistriFusion \cite{li2024distrifusion} proposes displaced patch parallelism, which splits the input image into patches and computes the noise for each patch in parallel, with the results from previous steps in replace of heavy communication to maintain consistency while minimizing overhead. And in AsyncDiff \cite{chen2024asyncdiff}, different components of the noise predicting model are distributed across different devices, where a warm-up phase and an asynchronous denoising scheme jointly enable its parallel execution. In contrast, mathematical methods often leverage the ODE/SDE formulation of the diffusion process and design alternative solvers with intrinsic parallelism. For instance, ParaDiGMS \cite{shih2023parallel} employs Picard iteration to solve the ODE so that parallelize sampling by predicting future denoising steps and iteratively refining them until convergence, enabling multiple diffusion steps to be denoised simultaneously. Important to note, this categorization does not provide a clear delineation but create a more clear landscape of the parallelism world. In our work, we combine a straightforward skip transition from the theoretical side and the system-level parallel implementation together, adopting both their advantages.
\section{Skip Transitions}

\begin{figure*}[t]
  \centering
  \captionsetup[subfigure]{justification=centering}

  \begin{subfigure}[t]{0.32\textwidth}
    \centering
    \includegraphics[width=\linewidth]{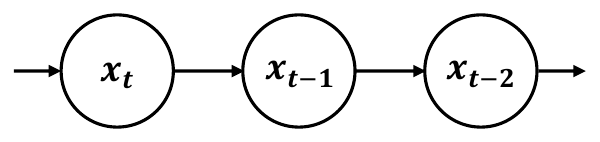}
    \caption{DDPM}\label{fig:ddpm_trans}
  \end{subfigure}\hfill
  \begin{subfigure}[t]{0.32\textwidth}
    \centering
    \includegraphics[width=\linewidth]{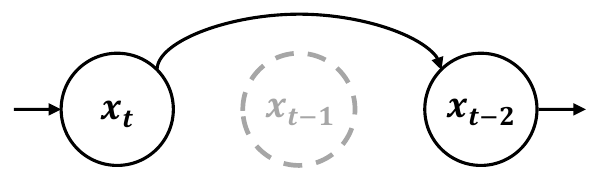}
    \caption{DDIM and Euler Solvers}\label{fig:ddim_trans}
  \end{subfigure}\hfill
  \begin{subfigure}[t]{0.32\textwidth}
    \centering
    \includegraphics[width=\linewidth]{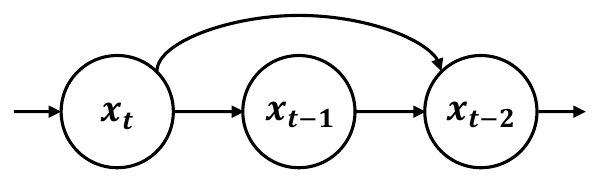}
    \caption{skip transition}
  \end{subfigure}

  \caption{Temporal dependencies of different methods. In DDPM, timesteps are sequential. DDIM and Euler Solvers allow global re-selection, but timestep traverse pattern within the subsequence or discretization remains sequential. The skip transitions allow direct moves between arbitrary pairs of timesteps, offering local temporal flexibility and enabling parallelization.}
  \label{fig:all_connection}
\end{figure*}

Skip transitions refer to predicting a future (or earlier) diffusion state $x_{t-k}$ directly from $(x_t,\varepsilon_t)$ by skipping intermediate timesteps. From a continuous-time perspective, skipping timesteps is a natural operation, since the system dynamics can be integrated over extended intervals. However, existing frameworks typically exercise this sequential freedom globally. For ODEs, it is dictated by the global discretization schedule, while for DDIM, it manifests as the re-selection of a timestep subsequence. To employ our Draft-and-Refine pattern, we need to \textbf{operatorize} this action, i.e., formally define this local skipping mechanism as an operator. This requires a derivation analogous to the backward formulas of each respective framework. In this section, we provide three examples: the classical DDPM \cite{ho2020denoising}, alongside the widely adopted DDIM \cite{song2020denoising} and Euler ODE solvers \cite{karras2022elucidating}.

\subsection{DDPM}
In the reverse process, DDPM denoises by \textbf{iteratively computing} $\boldsymbol{q_\theta(x_{t-1}|x_t)}$. At each step, we first estimate $\hat\varepsilon_t$ using the noise predictor, making our target to compute $q(x_{t-1}|x_t,\varepsilon_t)$, or equivalently $q(x_{t-1}|x_t,x_0)$, where $x_0$ is given by $x_t=\sqrt{\alpha_t} x_0 + \sqrt{1 - \alpha_t} \hat\varepsilon_t$. This target is tractable due to the availability of Bayes' rule, and this property holds even if we change it to the skip transition target $q(x_{t-k}|x_t,x_0)$, which would be:
\begin{equation}
    \label{eq:ddpm_bayes}
    q\left(x_{t-k} | x_t, x_0\right)
    = \frac{q\left(x_t | x_{t-k}\right)\ q\left(x_{t-k} | x_0\right)}{q\left(x_t | x_0\right)}.
\end{equation}
Since all terms on the RHS are Gaussian distributions, we can apply the same algebraic manipulations as in the original DDPM to obtain the closed-form skip transition:
\begin{equation}
    \label{eq:ddpm_skip_reverse}
    x_{t-k} | x_t, x_0 \sim \mathcal{N}\left(\mu'(x_t, x_0),\ \sigma_{t,k}^2\mathbf{I} \right),
\end{equation}
\begin{align}
    \label{eq:ddpm_skip_reverse_detail}
    \begin{gathered}
        \mu' =\frac{\sqrt{\frac{\alpha_t}{\alpha_{t-k}}}\left(1-\alpha_{t-k}\right) x_t+\sqrt{\alpha_{t-k}}\left(1-\frac{\alpha_t}{\alpha_{t-k}}\right) x_0}{1-\alpha_t}, \\
        \sigma_{t,k}^2 = \frac{\left(1-\frac{\alpha_t}{\alpha_{t-k}}\right)\left(1-\alpha_{t-k}\right)}{1-\alpha_t}.
    \end{gathered}
\end{align}
This result shows that the update is the k-step analogue of the one-step rule, following our intuition. Full steps for the derivation are detailed in Section \ref{sec:derivation}.

\subsection{DDIM}
Unlike DDPM, the diffusion process in DDIM is non-Markovian. Therefore, its denoising steps stem not from Bayes' rule, but rather from marginal consistency, which means composing the forward marginal $p(x_t|x_0)$ with the reverse transition $p(x_{t-1}|x_t)$ must strictly yield the marginal distribution $p(x_{t-1}|x_0)$. For a k-step skip transition, the consistency still holds, formally as:
\begin{equation}
    \label{eq:ddim_consistency}
    p(x_{t-k}|x_0) = \int p(x_{t-k}|x_t)\, p(x_t|x_0)\, \mathrm{d}x_t.
\end{equation}
Starting from this constraint, we can derive the result:
\begin{equation}
    \label{eq:ddim_skip_reverse}
    x_{t-k} \sim \mathcal{N}\left(\sqrt{\alpha_{t-k}}x_0 + \sqrt{1-\alpha_{t-k}-\sigma_{t,k}^2}\hat\varepsilon_t,\ \sigma_{t,k}^2\mathbf{I}\right),
\end{equation}
where $\hat\varepsilon_t = \varepsilon_\theta(x_t, t)$. As in the DDPM case, the update is still a k-step analogue, and full derivation is in Section \ref{sec:derivation}. The variance term, however, slightly differs in interpretation: in this notation, the usual DDIM $\sigma_t$ corresponds to $\sigma_{t,1}$, so all $\sigma_{t,k}$ where $k > 1$ are additional hyperparameters. In practice, DDIM typically employs zero variance (deterministic sampling), or alternatively, adopts the variance induced by DDPM; both choices carry over naturally to the skipped setting without ad-hoc tuning. Consequently, this skip transition can be seamlessly applied.

\subsection{Euler ODE Solvers}
The diffusion process can also be modeled as an ODE:
\begin{equation}
    \frac{\text{d}x}{\text{d}\sigma} = f(x, \sigma),
\end{equation}
where $\sigma$ is the continuous analog to the discrete timestep $t$, and $f(x, \sigma)$ dictates the system dynamics. For instance, $f(x, \sigma)$ can be the model-predicted velocity $v_\theta(x, \sigma)$ in Flow Matching \cite{lipman2022flow}, or $(x - \hat{x}_0)/\sigma$ in the original EDM \cite{karras2022elucidating}. For notational simplicity, we will use $v_\theta(x, \sigma)$ for illustration below. Applying first-order Euler method yields:
\begin{equation}
    \Delta x = \Delta\sigma\cdot v_\theta(x, \sigma).
\end{equation}
Since the ODE formulation operates in continuous time, skipping intermediate steps is mathematically equivalent to taking a larger numerical integration step. This allows us to handle skip transitions identically to standard single-step updates. Given a discretization schedule $\{\sigma_t\}_{t=1}^T$, transition from step $t$ to $t-k$ is straightforwardly given by:
\begin{equation}
    x_{t-k} = x_t + (\sigma_{t-k} - \sigma_t)\cdot v_\theta(x_t, \sigma_t),
\end{equation}

\paragraph{Conclusion} While the skip transition is a mathematically existing concept, we explicitly formulate it as an operator within the original frameworks without altering their underlying mechanics, providing a convenient primitive for subsequent parallelization. As illustrated in Figure~\ref{fig:all_connection}, our derivations allow us to connect any two diffusion states without redeciding the global timestep schedule. Consequently, this skip transition acts as a local operator that can be invoked on demand, significantly enhancing the flexibility of sampling pattern design and unlocking the potential for parallelization, as detailed in the next section.

\begin{figure*}[t]
  \centering
  \captionsetup[subfigure]{justification=centering}

  \hspace*{\fill}%
  \begin{subfigure}[t]{0.45\textwidth}
    \centering
    \includegraphics[width=\linewidth]{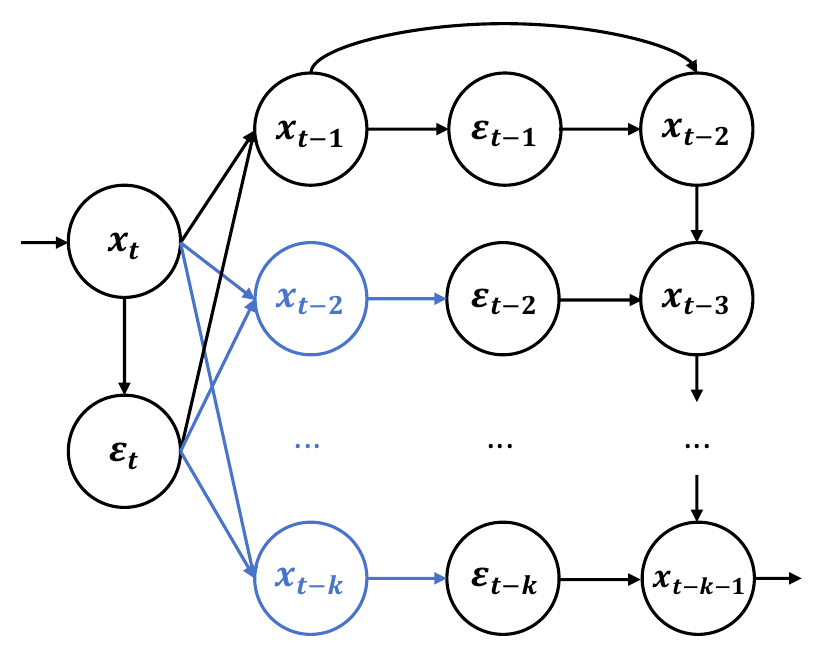}
    \caption{Computation map of conservative version of DRiffusion.}
    \label{fig:consv_map}
  \end{subfigure}\hfill
  \begin{subfigure}[t]{0.45\textwidth}
    \centering
    \includegraphics[width=\linewidth]{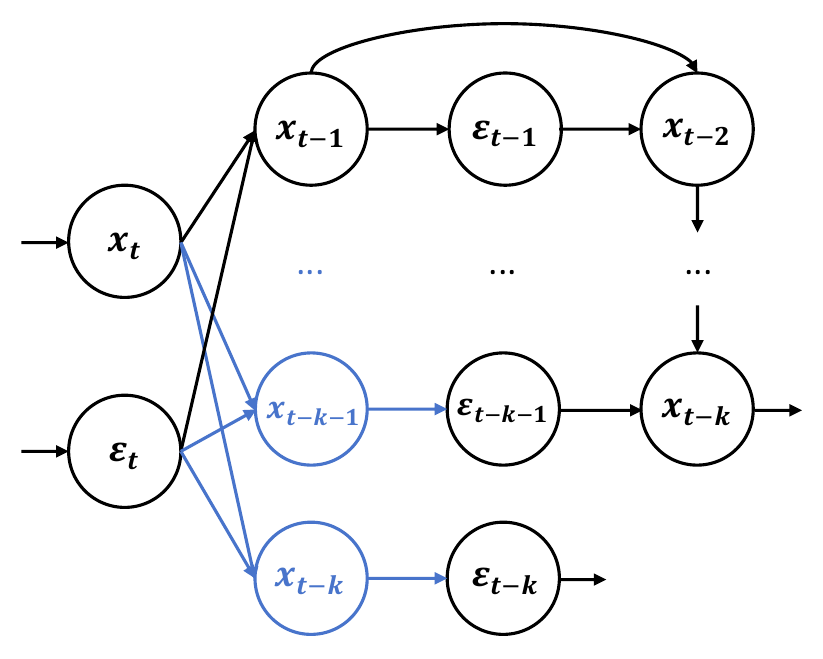}
    \caption{Computation map of aggressive version of DRiffusion.}
    \label{fig:aggr_map}
  \end{subfigure}
  \hspace*{\fill}%

  \caption{Computation map of different versions. Conservative one takes a standalone timestep for computing the noise for extra steps predicting while aggressive one directly takes the noise computed from draft states.}
  \label{fig:triple}
\end{figure*}

\section{DRiffusion}
A natural starting point for parallelizing diffusion models is to ask whether noise predictions at multiple timesteps can be computed simultaneously. In the original computation pipeline, this objective seems infeasible, since it requires access to all corresponding states at those timesteps without fully executing the denoising trajectory step by step. However, the perspective of skip transitions makes it possible: when each skip is treated as a local, callable operator, the required intermediate states can be constructed directly, thereby enabling parallel computation across timesteps.

Following this intuition, we leverage the flexibility to generate multiple states at subsequent steps concurrently, producing draft estimates of $x_{t-k}$ for several values of $k$ that remain aligned with the underlying denoising trajectory, albeit slightly less precise due to the larger step size. We then feed these drafts into the noise predictor to obtain reasonable noise estimates and proceed with the standard denoising updates to get refined states. In effect, the drafts serve as parallel proposals that are subsequently refined, unlocking substantial temporal acceleration while preserving the original structure of the underlying denoising trajectory.

Although previous works \cite{song2020denoising} have shown that very large step sizes degrade image quality due to imperfect noise prediction in practice, our empirical observations suggest two mitigating factors. First, mild perceptual degradation does not imply poor representation: the images or latent vectors often retain most of the underlying semantic and structural information. Second, even though the noise predictor is not exact, it generalizes well enough to map neighborhoods around plausible samples to reasonable outcomes. Building on these observations, we see that even when the draft-and-refine procedure uses larger step sizes, our DRiffusion method can still deliver sufficiently high-quality results.

\begin{algorithm*}[t]
\caption{Parallelization (Aggressive Version)}
\label{algo:aggr_para}
\begin{algorithmic}[1]
\Require sampled noise $x_T$, total steps $T$, block size $k$
\Ensure generated image $x_0$
\State $t \gets T$
\State $\varepsilon_t \gets \varepsilon_\theta(x_t, t)$
\Repeat
  \For{$i = 1$ to $k$ \textbf{concurrently}} \Comment{the following $k$ noise prediction can run in parallel}
    \State sample $z \sim \mathcal{N}(0,1)$
    \State $x_{t-i} \gets \sqrt{\alpha_{t-i}}\cdot
      \dfrac{x_t - \sqrt{1-\alpha_t}\cdot \varepsilon_t}{\sqrt{\alpha_t}}
      + \sqrt{1-\alpha_{t-i}-\sigma_{t,i}^2}\cdot \varepsilon_t +\sigma_{t,i}\cdot z$ 
    \State $\varepsilon_{t-i} \gets \varepsilon_\theta(x_{t-i}, t-i)$ 
    \Comment{computation bottleneck}
  \EndFor
  \For{$i = 2$ to $k$}
    \State sample $z \sim \mathcal{N}(0,1)$
    \State $x_{t-i} \gets \sqrt{\alpha_{t-i}}\cdot
      \dfrac{x_{t-i+1} - \sqrt{1-\alpha_{t-i+1}}\cdot \varepsilon_{t-i+1}}{\sqrt{\alpha_{t-i+1}}}
      + \sqrt{1-\alpha_{t-i}-\sigma_{t-i+1}^2}\cdot \varepsilon_{t-i+1}+\sigma_{t-i+1}\cdot z$
  \EndFor
  \State $t \gets t - k$
  \State $\text{cache }\varepsilon_{t-k}$ \Comment{$\epsilon_{t-k}$ was computed in parallel but it has not been used}
\Until{$t \le 0$}
\State \Return $x_0$
\end{algorithmic}
\end{algorithm*}

Algorithm~\ref{algo:aggr_para} presents the aggressive version of our method, using DDIM update as example. At current anchor timestep $t$, given state-noise pair $(x_t,\varepsilon_t)$, we use skip transitions to generate $k$ provisional drafts $x_{t-1}^{d},x_{t-2}^{d},\dots,x_{t-k}^{d}$. We then evaluate the network in parallel at timesteps $t-1,\dots,t-k$, taking these drafts as inputs to obtain the corresponding noise predictions $\varepsilon_{t-1},\varepsilon_{t-2},\dots,\varepsilon_{t-k}$. Using these predictions, we can perform the usual denoising updates to refine each state and, in particular, obtain a refined $x_{t-k}$ paired with $\varepsilon_{t-k}$ that becomes the next anchor. The $\varepsilon_{t-k}$ is computed during the parallelization, but not used in the updates. Therefore, we cache and carry it forward alongside $x_{t-k}$ to avoid an extra evaluation at the start of the next round, thereby saving one neural network forward pass per block and maximizing the overall sampling throughput.. Initialization requires a single evaluation at $t=T$ to obtain $\varepsilon_T$; thereafter the loop iterates until $x_0$ is produced. 

Because the $k$ noise predictions per iteration are fully parallelized in this scheme, this aggressive version admits an ideal wall-clock speedup of up to $k\times$ (i.e., runtime reduced by a factor of $1/k$), ignoring communication and other minor overheads. Figure \ref{fig:aggr_map} illustrates the computation map of our aggressive method, with blue elements denoting draft-related computations and states.

However, in this process, not all predicted noises sit in symmetric positions: the one used for computing multiple drafts seems to be more demanding than others, since larger step sizes are covered by this noise. To coordinate with this situation, we further propose the conservative version of our method. In this version, at each anchor timestep $t$, we only have current state $x_t$, and run a stand-alone noise prediction to get current noise. This noise is more accurate, since it is produced from the refined $x_t$ instead of the drafted ones, and with it we can replicate the same computation process in aggressive version, and even use $\varepsilon_{t-k}$ to push $x_{t-k}$ a step further to $x_{t-k-1}$ since we do not need to cache it for next iteration. This version of method has a speedup of up to $\frac{k+1}{2} \times$, as the anchor noise is predicted independently and each iteration pushes one step further. Figure \ref{fig:consv_map} illustrates the computation map of our conservative method, and see Algorithm~\ref{algo:consv_para} in Appendix for detailed information.

\begin{figure*}[t]
  \centering
  \setlength{\tabcolsep}{4pt}           
  \newcommand{\colpad}{0.5\fill}        
  \newcommand{\tilew}{0.235\textwidth}  

  \begin{tabular}{@{\hspace*{\colpad}} c c c c @{\hspace*{\colpad}}}

    \begin{minipage}[t]{\tilew}\centering
      \scriptsize \textbf{Original SD-2.1}\\[1pt] 
      \includegraphics[width=\linewidth]{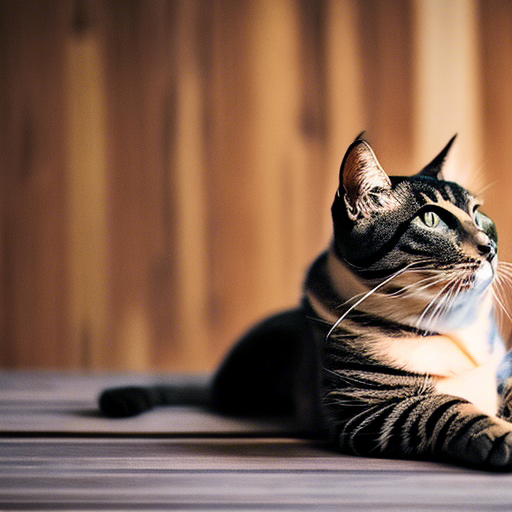}
    \end{minipage} &
    \begin{minipage}[t]{\tilew}\centering
      \scriptsize \textbf{N=2, Conservative, \textcolor{red}{1.5$\times$} Speedup}\\[1pt]
      \includegraphics[width=\linewidth]{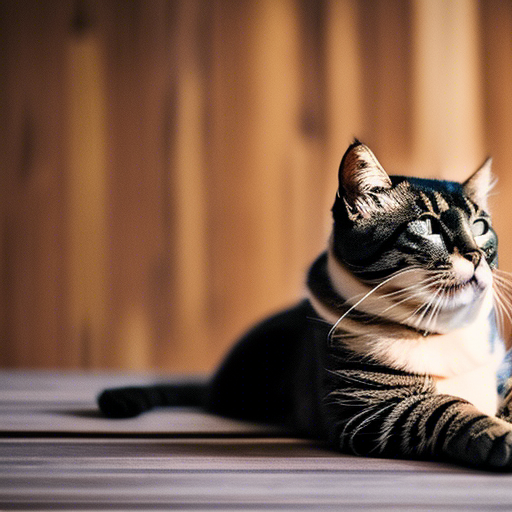}
    \end{minipage} &
    \begin{minipage}[t]{\tilew}\centering
      \scriptsize \textbf{N=3, Conservative, \textcolor{red}{1.9$\times$} Speedup}\\[1pt]
      \includegraphics[width=\linewidth]{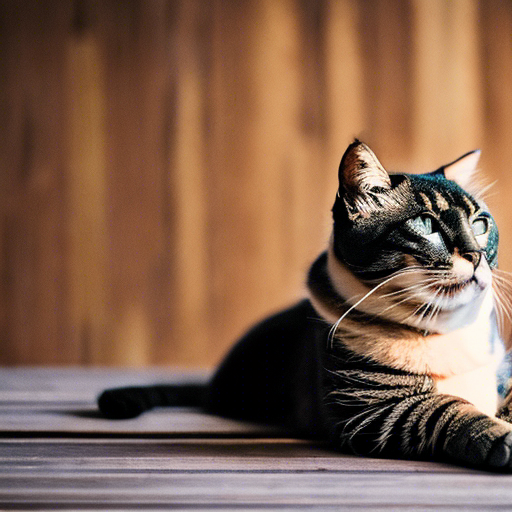}
    \end{minipage} &
    \begin{minipage}[t]{\tilew}\centering
      \scriptsize \textbf{N=4, Conservative, \textcolor{red}{2.5$\times$} Speedup}\\[1pt]
      \includegraphics[width=\linewidth]{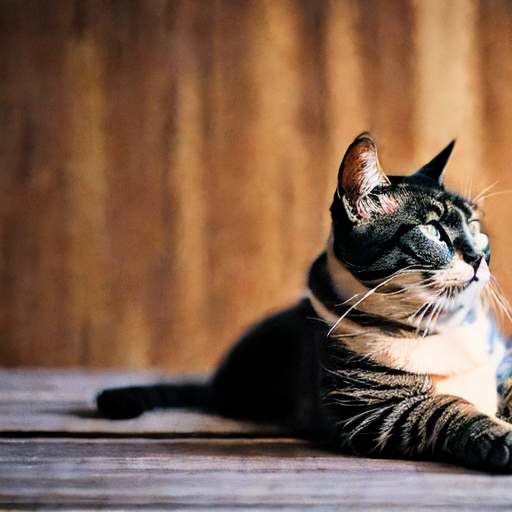}
    \end{minipage}
    \\ 

    \begin{minipage}[t]{\tilew}\centering
      \scriptsize \textbf{Original SD-2.1}\\[1pt]
      \includegraphics[width=\linewidth]{fig/grid_1_orig.png}
    \end{minipage} &
    \begin{minipage}[t]{\tilew}\centering
      \scriptsize \textbf{N=2, Aggressive, \textcolor{red}{1.9$\times$} Speedup}\\[1pt]
      \includegraphics[width=\linewidth]{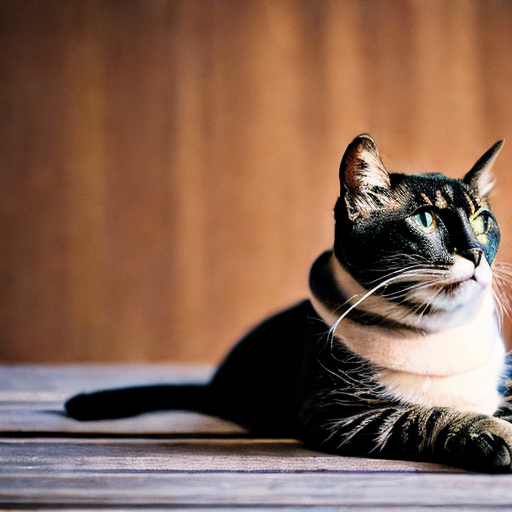}
    \end{minipage} &
    \begin{minipage}[t]{\tilew}\centering
      \scriptsize \textbf{N=3, Aggressive, \textcolor{red}{2.8$\times$} Speedup}\\[1pt]
      \includegraphics[width=\linewidth]{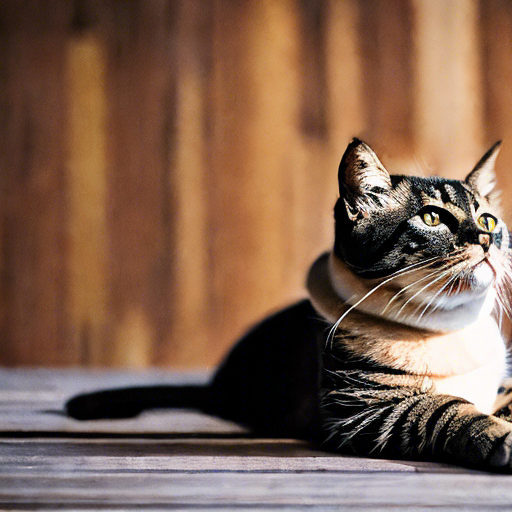}
    \end{minipage} &
    \begin{minipage}[t]{\tilew}\centering
      \scriptsize \textbf{N=4, Aggressive, \textcolor{red}{3.6$\times$} Speedup}\\[1pt]
      \includegraphics[width=\linewidth]{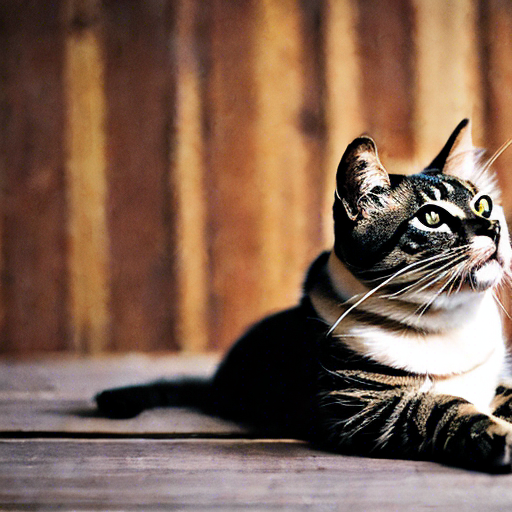}
    \end{minipage}

  \end{tabular}

  \caption{Qualitative comparison at 50 steps on MS-COCO using Stable Diffusion 2.1. The first row displays conservative results and the second row shows aggressive ones. Our parallelization method accelerates the sampling process with minimal impact on sample quality. The prompt used for generation is: \textit{A photo of a cat sitting on a wooden table in sunlight.}}
  \label{fig:qual_consv_aggr}
\end{figure*}

\section{Experiments}
\subsection{Setups}
\label{sec:setup}
\paragraph{Models and Datasets} We evaluate our method across a wide range of diffusion models, including Stable Diffusion 2.1 (SD2.1) \cite{rombach2022high}, a relatively small U-Net-based model; Stable Diffusion XL (SDXL) \cite{podell2023sdxl}, a larger U-Net-based model; and Stable Diffusion 3 (SD3) \cite{esser2024scaling}, a Transformer-based flow-matching model. Results on additional models, including Stable Diffusion 1.5 and Stable Diffusion 2.1 base \cite{rombach2022high}, are deferred to Section \ref{sec:quant_other}. This model coverage enables fair comparisons with prior methods while also demonstrating the adaptability of our approach across diverse architectures and model sizes. Unless otherwise specified, the number of samples per generation is set to one, and the number of inference steps follows the default setting of each model (28 steps for SD3, and 50 steps for the others). For benchmarking, we use the MS-COCO 2017 validation set \cite{lin2014microsoft}, which contains 5,000 images, each paired with five captions. Following common practice, we use only the first caption of each image for generation and text-image alignment evaluation to ensure a one-to-one correspondence between generated images and reference captions.

\vspace{-3pt}
\paragraph{Evaluation Metrics} Consistent with prior research, our primary metrics for assessing perceptual generation quality are the Fréchet Inception Distance (FID)~\cite{heusel2017gans}, computed on Inception-V3 features~\cite{szegedy2016rethinking}, and CLIPScore (CLIP)~\cite{radford2021learning, hessel2021clipscore}, evaluated using the \textit{openai/clip-vit-large-patch14} model, which provides more robust measurements than the \textit{openai/clip-vit-base-patch32} one. However, since these traditional metrics can lack sensitivity to fine-grained visual preferences, we supplement our evaluation with the more recent PickScore (Pick)~\cite{kirstain2023pick} and Human Preference Score v2.1 (HPSv2.1)~\cite{wu2023human} in certain cases. For efficiency evaluation, we measure the average sampling latency across multiple steady-state runs using up to four NVIDIA V100 GPUs. We report the relative speedup compared to a single-GPU diffusion baseline, along with the extra memory overhead (ExtraMem.) incurred by our method.

\vspace{-3pt}
\paragraph{Baselines} We compare DRiffusion against two representative acceleration approaches:
(i) direct skip transitions, instantiated as reduced sampling steps, and
(ii) AsyncDiff~\cite{chen2024asyncdiff}, which parallelizes denoising by distributing sub-networks across multiple devices and performing asynchronous sampling. To ensure a unified evaluation, we reproduced the AsyncDiff results using its official implementation under our measurement settings.

\subsection{Qualitative Result}
In Figure \ref{fig:qual_consv_aggr}, we present a qualitative comparison of the generated images. Based on Stable Diffusion 2.1 base model, we provide results using both the conservative and the aggressive acceleration strategies, along with the corresponding speedup ratios, where N denotes the number of GPUs used. Despite being inherently difficult to achieve outputs that are visually identical with the baseline at high speedup ratios, our method consistently produces images that preserve the semantic content of the text prompt. This can be observed in fine details such as the texture of the wooden table and the sunlight highlights on the cat’s chest.

Interestingly, by moderately skipping parts of the noise sampling process and reusing latent features, the accelerated variants may occasionally exhibit stronger local contrast or sharper details in certain regions than the baseline. For example, the reflections in the cat’s eyes can appear more pronounced. Nevertheless, aggressive acceleration may also introduce a slight drop in image fidelity. At speedup ratios approaching $4\times$, the generated images may show oversaturated colors or subtle artifacts, leading to a less clean appearance while still remaining globally consistent with the baseline. Overall, these results demonstrate that our method preserves high generation quality across a wide range of acceleration levels.

\begin{table*}[t]
    \centering
    \caption{Quantitative results of DRiffusion under different models on the MS-COCO dataset. Latency is averaged over steady runs. Speedup is measured against the 1-device original run of the same model. Number of samples per generation is 1.}
    \label{tab:full}
    \small
    \begin{tabularx}{0.9\linewidth}{cccCCCCCC}
        \toprule
        \textbf{Model} & \textbf{Mode} & \textbf{\#Devices} & \textbf{Latency$\ $(s)} & \textbf{Speed Up} & \textbf{FID} $\downarrow$ & \textbf{CLIP} $\uparrow$ & \textbf{Pick} $\uparrow$ & \textbf{HPSv2.1}$\ \uparrow$ \\
        \midrule
        \multirow{7}{*}{\textbf{SD2.1}}
          & original & 1 & 5.96 & -- & 23.63 & 26.29 & 21.79 & 26.62 \\
        \cmidrule(l){2-9}
          & \multirow{3}{*}{conservative}
              & 2 & 4.09 & 1.5$\times$ & 23.60 & 26.32 & 21.78 & 26.61 \\
          &   & 3 & 3.12 & 1.9$\times$ & 23.50 & 26.32 & 21.77 & 26.56 \\
          &   & 4 & 2.42 & 2.5$\times$ & 23.47 & 26.30 & 21.75 & 26.49 \\
        \cmidrule(l){2-9}
          & \multirow{3}{*}{aggressive}
              & 2 & 3.06 & 1.9$\times$ & 23.39 & 26.31 & 21.77 & 26.56 \\
          &   & 3 & 2.11 & 2.8$\times$ & 23.26 & 26.29 & 21.73 & 26.47 \\
          &   & 4 & 1.66 & 3.6$\times$ & 23.24 & 26.27 & 21.69 & 26.34 \\
        \midrule
        \multirow{7}{*}{\textbf{SDXL}}
          & original & 1 & 13.23 & -- & 24.02 & 26.67 & 22.43 & 27.50 \\
        \cmidrule(l){2-9}
          & \multirow{3}{*}{conservative}
              & 2 & 9.14 & 1.4$\times$ & 23.87 & 26.60 & 22.37 & 27.30 \\
          &   & 3 & 6.99 & 1.9$\times$ & 23.87 & 26.56 & 22.36 & 27.23 \\
          &   & 4 & 5.38 & 2.5$\times$ & 23.88 & 26.51 & 22.33 & 27.07 \\
        \cmidrule(l){2-9}
          & \multirow{3}{*}{aggressive}
              & 2 & 6.86 & 1.9$\times$ & 24.08 & 26.61 & 22.33 & 27.13 \\
          &   & 3 & 4.72 & 2.8$\times$ & 24.11 & 26.59 & 22.30 & 27.00 \\
          &   & 4 & 3.65 & 3.6$\times$ & 24.14 & 26.54 & 22.26 & 26.84 \\
        \midrule
        \multirow{7}{*}{\textbf{SD3}}
          & original & 1 & 11.25 & -- & 31.03 & 26.56 & 22.57 & 29.14 \\
        \cmidrule(l){2-9}
          & \multirow{3}{*}{conservative}
              & 2 & 7.60 & 1.5$\times$ & 31.80 & 26.54 & 22.44 & 28.82 \\
          &   & 3 & 5.49 & 2.0$\times$ & 30.82 & 26.51 & 22.29 & 28.56 \\
          &   & 4 & 4.63 & 2.4$\times$ & 29.94 & 26.55 & 22.16 & 28.32 \\
        \cmidrule(l){2-9}
          & \multirow{3}{*}{aggressive}
              & 2 & 5.89 & 1.9$\times$ & 31.32 & 26.52 & 22.34 & 28.66 \\
          &   & 3 & 3.81 & 3.0$\times$ & 29.73 & 26.50 & 22.12 & 28.20 \\
          &   & 4 & 3.00 & 3.7$\times$ & 29.08 & 26.46 & 21.95 & 27.64 \\
        \bottomrule
    \end{tabularx}
\end{table*}

\subsection{Quantitative Result}
Table~\ref{tab:full} reports the quality and efficiency of DRiffusion under different configurations on MS-COCO across three distinct generation models: SD2.1, SDXL, and SD3. We organize our discussion into two key aspects:

\paragraph{Generation Quality}
Across all tested settings, DRiffusion maintains a generation fidelity that is highly consistent with the original single-device baselines. FID generally fluctuates at a comparable level, while the CLIP score drops by at most 0.16 across all cases. In some circumstances, DRiffusion even yields slightly better FID scores, suggesting that introducing parallel draft states does not compromise semantic alignment or visual realism. However, we attribute these minor gains to statistical variance and the inherent sensitivity of FID when evaluating subtle variations in our cases, rather than to a genuine improvement.

To provide a more robust assessment, we further evaluate the method using PickScore (Pick) and Human Preference Score v2.1 (HPSv2.1). DRiffusion consistently delivers results closely aligned with the original models, evidenced by a marginal average drop of only 0.17 for Pick and 0.43 for HPSv2.1 across all configurations. It is worth noting one exception: SD3 in aggressive mode with 4 devices, where HPSv2.1 drops by 1.50. This steeper decline occurs because SD3 defaults to a much smaller number of sampling steps (only 28); consequently, applying the aggressive method with the largest step size could amplify the approximation error. Nevertheless, because each metric possesses its own sensitivities, a holistic view of the combined stability across all four metrics, alongside the substantial inference acceleration, confirms that this remains an efficient and reasonable trade-off. To conclude, our method incurs a negligible performance drop in the tested cases while achieving significant wall-clock speedups.

\begin{figure}[t]
  \centering
  \includegraphics[width=\linewidth]{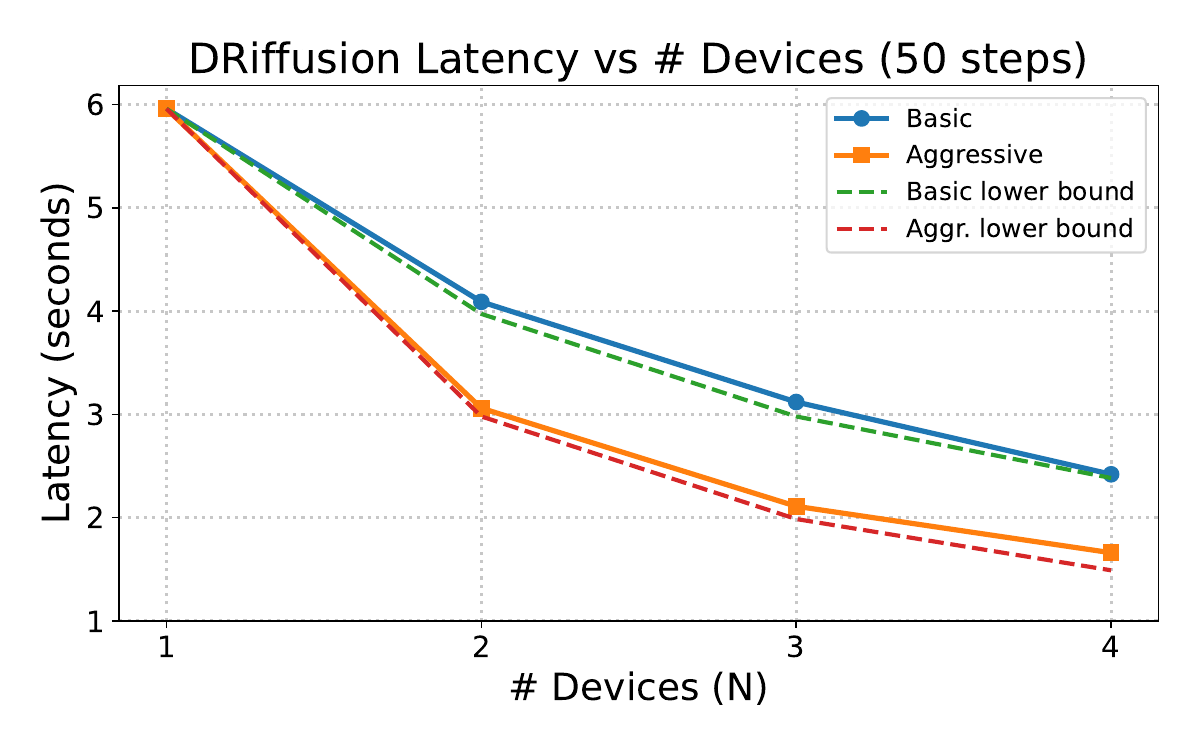}
  \caption{Latency scaling of DRiffusion with respect to the number of devices (N) on Stable Diffusion 2.1 base using 50 sampling steps. Both modes closely follow their theoretical lower bounds. }
  \label{fig:latency}
  \vspace{-1em}
\end{figure}

\paragraph{Inference Acceleration} We report sampling latencies measured on the denoising iterations under different configuration in Table~\ref{tab:full}, excluding fixed overheads. As presented, DRiffusion achieves substantial latency reduction that scales favorably with the number of computational devices. The speedup ranges from $1.4\times$ to $3.7\times$ depending on the mode and device count, while maintaining nearly the same total computation per sample. Fig.~\ref{fig:latency} shows that the aggressive mode approaches its theoretical lower bound of $\mathcal{O}(1/N)$ latency scaling, whereas the conservative mode closely tracks the expected $\mathcal{O}(2/(N+1))$ behavior. These observations demonstrate that DRiffusion provides efficient and scalable parallelism, substantially improving responsiveness for real-time or interaction-intensive applications.

\begin{table*}[t]
    \centering
    \caption{Comparison with baselines on MS-COCO dataset using SDXL. Speedups and ExtraMems are measured relatively to the 1-device SDXL baseline, and one sample per generation. We set the number of warm-up steps of AysncDiff to 5 to match our acceleration levels.}
    \label{tab:comparison}
    \small
    \begin{tabularx}{0.88\linewidth}{l C C CCCC}
        \toprule
        \textbf{Method} &
        \textbf{Devices} &
        \textbf{SpeedUp}$\uparrow$ &
        \textbf{ExtraMem.}$\downarrow$ &
        \textbf{FID}$\downarrow$ &
        \textbf{CLIP}$\uparrow$ &
        \textbf{Pick}$\uparrow$ \\
        \midrule
        Stable Diffusion & 1 & -- & +0MB  & 24.02 & 26.67 & 22.43 \\
        \midrule
        AsyncDiff (N=2, S=1) & 2 & 1.6$\times$ & +494MB & 24.13 & \textbf{26.60} & 22.34 \\
        DRiffusion (consv., 2 dev) & 2 & 1.5$\times$ & +186MB & \textbf{23.87} & \textbf{26.60} & \textbf{22.37} \\
        \midrule
        Stable Diffusion (step=25) & 1 & 2.0$\times$ & +0MB & 24.51 & \textbf{26.64} & 22.27 \\
        AsyncDiff (N=3, S=1) & 3 & 2.1$\times$ & +502MB & 24.17 & 26.56 & 22.30 \\
        DRiffusion (aggr., 2 dev) & 2 & 1.9$\times$ & +186MB & 24.08 & 26.61 & 22.33 \\
        DRiffusion (consv., 3 dev) & 3 & 1.9$\times$ & +220MB & \textbf{23.87} & 26.56 & \textbf{22.36} \\
        \midrule
        AsyncDiff (N=4, S=1) & 4 & 2.5$\times$ & +574MB & 23.92 & 26.51 & 22.20 \\
        AsyncDiff (N=2, S=2) & 3 & 2.7$\times$ & +532MB & 23.98 & 26.58 & 22.13 \\
        DRiffusion (consv., 4 dev) & 4 & 2.4$\times$ & +226MB & \textbf{23.88} & 26.51 & \textbf{22.33} \\
        DRiffusion (aggr., 3 dev) & 3 & 2.8$\times$ & +198MB & 24.11 & \textbf{26.59} & 22.30 \\
        \midrule
        AsyncDiff (N=3, S=2) & 4 & 3.4$\times$ & +554MB & 24.25 & \textbf{26.54} & 22.02 \\
        DRiffusion (aggr., 4 dev) & 4 & 3.6$\times$ & +222MB & \textbf{24.14} & \textbf{26.54} & \textbf{22.26} \\
        \bottomrule
    \end{tabularx}
\end{table*}

\subsection{Comparison}
Table~\ref{tab:comparison} presents quantitative data comparing DRiffusion with AsyncDiff under Stable Diffusion XL on the MS-COCO dataset. To ensure a fair comparison, we evaluate AsyncDiff using its official codebase, compute metrics following standard practices as detailed in Section \ref{sec:setup}, and set its number of warm-up steps to 5 to match our acceleration levels. We group the runs by similar speedups, ranging from 1.5$\times$ to 3.6$\times$, to comprehensively demonstrate the performance across different acceleration regimes.

Across all speed up groups, DRiffusion consistently achieves better generative quality relative to both AsyncDiff and the reduced-step SDXL. While FID and CLIP scores exhibit statistical fluctuations, DRiffusion still maintains a leading edge within the matched groups. And to properly account for these fluctuations, we further adopt PickScore as the more robust and sensitive assessment metric for these acceleration settings. Under PickScore, our method consistently outperforms the baselines by a substantial margin. Notably, across all evaluated acceleration regimes, DRiffusion reduces the PickScore performance degradation gap by an average of 48.6\%, achieving a maximum gap reduction of 58.5\% under the most extreme acceleration setting (4-device acceleration).

In terms of computational efficiency, DRiffusion scales nearly proportionally with the number of devices, demonstrating its effectiveness. When operating under similar device counts, our method closely matches or consistently surpasses the speedups of AsyncDiff. Moreover, DRiffusion demonstrates exceptional memory efficiency, a critical factor for high-resolution generation tasks. As shown in Table~\ref{tab:comparison}, AsyncDiff suffers from an escalating memory burden, requiring up to +574MB of extra memory as the device count and step sizes increase. In stark contrast, our method introduces a minimal and relatively stable overhead ranging from +186MB to +226MB across all multi-device configurations. This overhead is negligible, considering that the original SDXL baseline requires around 13004MB of memory. This advantage becomes even more pronounced when increasing the batch size. On a 32GB node, DRiffusion operates seamlessly with SDXL at a batch size of 5, while AsyncDiff encounters Out-Of-Memory (OOM) exceptions under the same setting. This is because our method modifies only the iterative sampling process, remaining independent of the model structure and the computations. By decoupling the memory cost from the acceleration regime, this low footprint prevents potential OOM bottlenecks, making DRiffusion more scalable and practical for deployment.

\section{Conclusion}

We introduced \emph{DRiffusion}, a draft-and-refine parallel sampling framework that unlocks inherent parallelism in diffusion inference via analytically valid \emph{skip transitions}. By drafting multiple future states and predicting their noises in parallel—then refining along the original denoising trajectory—DRiffusion consolidates the dominant network evaluations into a single parallel step without modifying the pretrained model or sampler. We instantiate two modes that offer a quality–speed trade-off: an \emph{aggressive} variant whose ideal latency scales as $\mathcal{O}(1/N)$ (up to $N\times$ speedup), and a \emph{conservative} variant with latency $\mathcal{O}\!\left(2/(N+1)\right)$ (up to $\tfrac{N+1}{2}\times$ speedup). across multiple diffusion models on MS-COCO, DRiffusion achieves $1.4\times$–$3.7\times$ wall-clock acceleration on up to four GPUs while keeping FID and CLIP essentially unchanged, with PickScore and HPSv2.1 remaining broadly comparable to the original sampler, at only minimal and controllable memory overhead. Compared with reduced-step sampling and system-level parallelization (e.g., AsyncDiff), DRiffusion consistently improves perceptual quality at similar acceleration rates, establishing a simpler, more powerful baseline for high-resolution and scalable parallel diffusion inference.

\paragraph{Acknowledgments.}
This work was supported in part by the China National Key R\&D Program under Grant No. 2023YFB3307201) and Putuo District (Shanghai) FuturististAI Lab Foundation under Grant No. QH2024-03-001.

{
    \small
    \bibliographystyle{ieeenat_fullname}
    \bibliography{main}
}

\clearpage
\setcounter{page}{1}
\maketitlesupplementary

\section{Derivation for Skip Transition}
\label{sec:derivation}
For DDPM:
\begin{align}
\begin{aligned} 
&q\left(\boldsymbol{x}_{t-k} | \boldsymbol{x}_t, \boldsymbol{x}_0\right) \\
=\,&\frac{q\left(\boldsymbol{x}_t | \boldsymbol{x}_{t-k}\right) q\left(\boldsymbol{x}_{t-k} | \boldsymbol{x}_0\right)}{q\left(\boldsymbol{x}_t | \boldsymbol{x}_0\right)} \\ =\,&\frac{\mathcal{N}\left(\boldsymbol{x}_t ; \sqrt{\prod_{i=t-k+1}^t\alpha_i} \boldsymbol{x}_{t-1},\left(1-\prod_{i=t-k+1}^t\alpha_i\right) \mathbf{I}\right) \mathcal{N}\left(\boldsymbol{x}_{t-k} ; \sqrt{\alpha_{t-k}} \boldsymbol{x}_0,\left(1-\alpha_{t-k}\right) \mathbf{I}\right)}{\mathcal{N}\left(\boldsymbol{x}_t ; \sqrt{\alpha_t} \boldsymbol{x}_0,\left(1-\alpha_t\right) \mathbf{I}\right)}\\ 
\propto\,&\exp \left\{-\frac{1}{2}\left[\frac{\left(\boldsymbol{x}_t-\sqrt{\frac{\alpha_t}{\alpha_{t-k}}} \boldsymbol{x}_{t-1}\right)^2}{1-\frac{\alpha_t}{\alpha_{t-k}}}+\frac{\left(\boldsymbol{x}_{t-k}-\sqrt{\alpha_{t-k}} \boldsymbol{x}_0\right)^2}{1-\alpha_{t-k}}-\frac{\left(\boldsymbol{x}_t-\sqrt{\alpha_t} \boldsymbol{x}_0\right)^2}{1-\alpha_t}\right]\right\}\\ 
\propto\,& \exp \left\{-\frac{1}{2}\left[-\frac{2 \sqrt{\frac{\alpha_t}{\alpha_{t-k}}} \boldsymbol{x}_t \boldsymbol{x}_{t-k}}{1-\frac{\alpha_t}{\alpha_{t-k}}}+ \frac{\frac{\alpha_t}{\alpha_{t-k}} \boldsymbol{x}_{t-k}^2}{1-\prod_{i=t-k+1}^t \alpha_i}+\frac{\boldsymbol{x}_{t-k}^2}{1-\alpha_{t-k}}-\frac{2 \sqrt{\alpha_{t-k}} \boldsymbol{x}_{t-k} \boldsymbol{x}_0}{1-\alpha_{t-k}}+C\left(\boldsymbol{x}_t, \boldsymbol{x}_0\right)\right]\right\} \\
\propto\,&\exp \left\{-\frac{1}{2}\left[\frac{1-\alpha_t}{\left(1-\frac{\alpha_t}{\alpha_{t-k}}\right)\left(1-\alpha_{t-k}\right)} \boldsymbol{x}_{t-k}^2-2\left(\frac{\sqrt{\frac{\alpha_t}{\alpha_{t-k}}} \boldsymbol{x}_t}{1-\frac{\alpha_t}{\alpha_{t-k}}}+\frac{\sqrt{\alpha_{t-k}} \boldsymbol{x}_0}{1-\alpha_{t-k}}\right) \boldsymbol{x}_{t-k}\right]\right\}\\ 
\propto\,&\exp \left\{-\frac{1}{2}\left(\frac{1-\alpha_t}{\left(1-\frac{\alpha_t}{\alpha_{t-k}}\right)\left(1-\alpha_{t-k}\right)}\right)\left[\boldsymbol{x}_{t-k}^2-2 \frac{\sqrt{\frac{\alpha_t}{\alpha_{t-k}}}\left(1-\alpha_{t-k}\right) \boldsymbol{x}_t+\sqrt{\alpha_{t-k}}\left(1-\frac{\alpha_t}{\alpha_{t-k}} \right) \boldsymbol{x}_0}{1-\alpha_t} \boldsymbol{x}_{t-k}\right]\right\}\\ 
=\,&\exp \left\{-\frac{\left(\boldsymbol{x}_{t-k} - \frac{\sqrt{\frac{\alpha_t}{\alpha_{t-k}}}\left(1-\alpha_{t-k}\right) \boldsymbol{x}_t+\sqrt{\alpha_{t-k}}\left(1-\frac{\alpha_t}{\alpha_{t-k}}\right) \boldsymbol{x}_0}{1-\alpha_t}\right)^2}{2 \left( \frac{\left(1-\frac{\alpha_t}{\alpha_{t-k}}\right)\left(1-\alpha_{t-k}\right)}{1-\alpha_t}\right)}\right\}\\ \propto\,& \mathcal{N}\left(\boldsymbol{x}_{t-k} ; \underbrace{\frac{\sqrt{\frac{\alpha_t}{\alpha_{t-k}}}\left(1-\alpha_{t-k}\right) \boldsymbol{x}_t+\sqrt{\alpha_{t-k}}\left(1-\frac{\alpha_t}{\alpha_{t-k}}\right) \boldsymbol{x}_0}{1-\alpha_t}}_{\mu_q\left(\boldsymbol{x}_t, \boldsymbol{x}_0\right)}, \underbrace{\frac{\left(1-\frac{\alpha_t}{\alpha_{t-k}}\right)\left(1-\alpha_{t-k}\right)}{1-\alpha_t}}_{\boldsymbol{\Sigma}_q(t)}\mathbf{I}\right).
\end{aligned}
\end{align}

\noindent
For DDIM, we assume skip transitions are Gaussian distribution with unknown parameters: \begin{equation}
p(\boldsymbol{x}_{t-1}|\boldsymbol{x}_t, \boldsymbol{x}_0) = \mathcal{N}(\boldsymbol{x}_{t-1}; \kappa_t \boldsymbol{x}_t + \lambda_t \boldsymbol{x}_0, \sigma_t^2 \boldsymbol{I}).
\end{equation}
Using forward equations, Marginal consistency Eq.~\eqref{eq:ddim_consistency} gives the constraints:
\begin{equation}
    \lambda_{t,t-k} + \kappa_{t,t-k}\sqrt{\alpha_t} = \sqrt{\alpha_{t-k}},\quad \kappa_{t,t-k}^2(1-\alpha_t)+\sigma_{t-k+1}^2 = 1 - \alpha_{t-k}.
\end{equation}
Solving, we obtain:
\begin{equation}
    \kappa_{t,t-k} = \frac{\sqrt{1-\alpha_{t-k}-\sigma^2}}{\sqrt{1-\alpha_t}},\quad \lambda_{t,t-k} = \sqrt{\alpha_{t-k}} - \frac{\sqrt{1-\alpha_{t-k}-\sigma^2}}{\sqrt{1-\alpha_t}}\sqrt{\alpha_t}.
\end{equation}
Substituting back yields
\begin{align}
    \begin{aligned}
    & P(x_{t-1}|x_t,x_0) \\
    =\, & \mathcal{N}(x_{t-1};(\sqrt{\alpha_{t-1}} - \frac{\sqrt{1-\alpha_{t-1}-\sigma^2}}{\sqrt{1-\alpha_t}}\sqrt{\alpha_t})x_0 + (\frac{\sqrt{1-\alpha_{t-1}-\sigma^2}}{\sqrt{1-\alpha_t}})x_t,\sigma_t^2) \\
    =\, & \mathcal{N}(x_{t-1};\sqrt{\alpha_{t-1}}x_0 + \sqrt{1-\alpha_{t-1}-\sigma^2}\epsilon_t,\sigma_t^2),
    \end{aligned}
\end{align}
and
\begin{equation}
    x_{t-k} = \sqrt{\alpha_{t-k}} \underbrace{\left( \frac{x_t - \sqrt{1-\alpha_t}\epsilon_\theta(x_t,t) }{\sqrt{\alpha_t}} \right)}_{\text{predicted }x_0} + \sqrt{1-\alpha_{t-k}-\sigma_t^2} \cdot \epsilon_\theta(x_t,t) + \sigma_tz_t.
\end{equation}

\section{Algorithm for Conservative Parallelization}
\begin{algorithm*}[!ht]
\caption{Parallelization (Conservative Version)}
\label{algo:consv_para}
\begin{algorithmic}[1]
\Require sampled noise $x_T$, total steps $T$, block size $k$
\Ensure generated image $x_0$
\State $t \gets T$
\Repeat
  \State $\varepsilon_t \gets \varepsilon_\theta(x_t, t)$
  \For{$i = 1$ to $k$ \textbf{concurrently}} \Comment{the following $k$ noise prediction can run in parallel}
    \State sample $z \sim \mathcal{N}(0,1)$
    \State $x_{t-i} \gets \sqrt{\alpha_{t-i}}\cdot
      \dfrac{x_t - \sqrt{1-\alpha_t}\cdot \varepsilon_t}{\sqrt{\alpha_t}}
      + \sqrt{1-\alpha_{t-i}-\sigma_{t,i}^2}\cdot \varepsilon_t +\sigma_{t,i}\cdot z$ 
    \State $\varepsilon_{t-i} \gets \varepsilon_\theta(x_{t-i}, t-i)$ \Comment{computation bottleneck}
  \EndFor
  \For{$i = 2$ to $k+1$}
    \State sample $z \sim \mathcal{N}(0,1)$
    \State $x_{t-i} \gets \sqrt{\alpha_{t-i}}\cdot
      \dfrac{x_{t-i+1} - \sqrt{1-\alpha_{t-i+1}}\cdot \varepsilon_{t-i+1}}{\sqrt{\alpha_{t-i+1}}}
      + \sqrt{1-\alpha_{t-i}-\sigma_{t-i+1}^2}\cdot \varepsilon_{t-i+1}+\sigma_{t-i+1}\cdot z$
  \EndFor
  \State $t \gets t - k - 1$
\Until{$t \le 0$}
\State \Return $x_0$
\end{algorithmic}
\end{algorithm*}

\section{Additional Experiments}
\subsection{Quatitative Results on Other Models}
\label{sec:quant_other}
We provide additional quantitative results on SD1.5 and SD2.1-base in Table~\ref{tab:other}. The overall trend is consistent with the main-text observations: DRiffusion continues to offer substantial wall-clock acceleration while maintaining quality comparable to the original sampler. On SD1.5, the method achieves up to 3.4$\times$ speedup, with FID, CLIP, Pick, and HPSv2.1 all remaining close to the original setting. A similar pattern is observed on SD2.1-base, where the speedup reaches 3.5$\times$. These additional results further confirm that the favorable efficiency--quality trade-off of DRiffusion generalizes across different diffusion backbones.

\begin{table*}[t]
    \centering
    \caption{Quantitative results of DRiffusion under SD1.5 and SD2.1-base on the MS-COCO dataset.}
    \label{tab:other}
    \small
    \begin{tabularx}{0.9\linewidth}{cccCCCCCC}
        \toprule
        \textbf{Model} & \textbf{Mode} & \textbf{\#Devices} & \textbf{Latency$\ $(s)} & \textbf{Speed Up} & \textbf{FID} $\downarrow$ & \textbf{CLIP} $\uparrow$ & \textbf{Pick} $\uparrow$ & \textbf{HPSv2.1}$\ \uparrow$ \\
        \midrule
        \multirow{7}{*}{\textbf{SD1.5}}
          & original & 1 & 2.68 & -- & 25.91 & 26.49 & 21.49 & 26.07 \\
        \cmidrule(l){2-9}
          & \multirow{3}{*}{conservative}
              & 2 & 2.07 & 1.3$\times$ & 25.95 & 26.60 & 21.51 & 25.98 \\
          &   & 3 & 1.43 & 1.9$\times$ & 25.67 & 26.60 & 21.51 & 25.95 \\
          &   & 4 & 1.10 & 2.4$\times$ & 25.57 & 26.61 & 21.50 & 25.92 \\
        \cmidrule(l){2-9}
          & \multirow{3}{*}{aggressive}
              & 2 & 1.41 & 1.9$\times$ & 26.04 & 26.55 & 21.50 & 25.94 \\
          &   & 3 & 1.00 & 2.7$\times$ & 25.48 & 26.58 & 21.49 & 25.90 \\
          &   & 4 & 0.80 & 3.4$\times$ & 25.36 & 26.55 & 21.45 & 25.81 \\
        \midrule
        \multirow{7}{*}{\textbf{\shortstack{SD2.1\\base}}}
          & original & 1 & 2.48 & -- & 25.69 & 26.19 & 21.81 & 27.14 \\
        \cmidrule(l){2-9}
          & \multirow{3}{*}{conservative}
              & 2 & 1.74 & 1.4$\times$ & 25.75 & 26.26 & 21.83 & 27.04 \\
          &   & 3 & 1.32 & 1.9$\times$ & 25.67 & 26.27 & 21.82 & 27.02 \\
          &   & 4 & 1.04 & 2.4$\times$ & 25.56 & 26.26 & 21.81 & 26.99 \\
        \cmidrule(l){2-9}
          & \multirow{3}{*}{aggressive}
              & 2 & 1.35 & 1.8$\times$ & 25.45 & 26.26 & 21.82 & 27.02 \\
          &   & 3 & 0.92 & 2.7$\times$ & 25.46 & 26.24 & 21.80 & 26.96 \\
          &   & 4 & 0.72 & 3.5$\times$ & 25.14 & 26.21 & 21.76 & 26.88 \\
        \bottomrule
    \end{tabularx}
\end{table*}

\subsection{Comparison with Distillation Method}
\label{sec:compare_distill}
\begin{figure}[b]
  \centering
  \vspace{-10pt}
  \setlength{\tabcolsep}{3pt}

  \begin{tabular}{@{} m{.08\linewidth} @ {} m{.19\linewidth} m{.19\linewidth} m{.19\linewidth} @{} m{.08\linewidth} @{}}
    & \centering \textit{Original} &
    \centering \textit{DRiffusion} &
    \centering \textit{LCM}  & 
    \\

    & \centering\includegraphics[width=\linewidth]{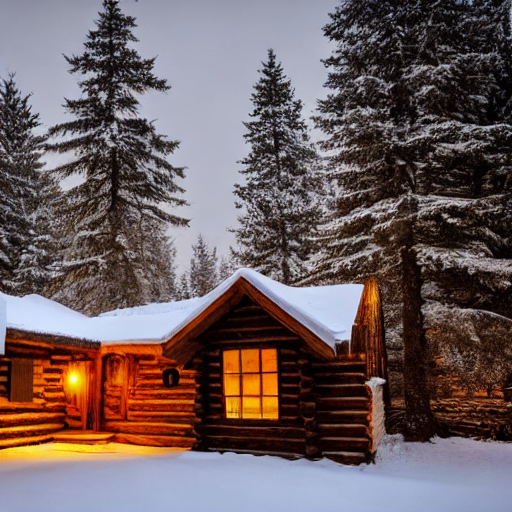} &
    \centering\includegraphics[width=\linewidth]{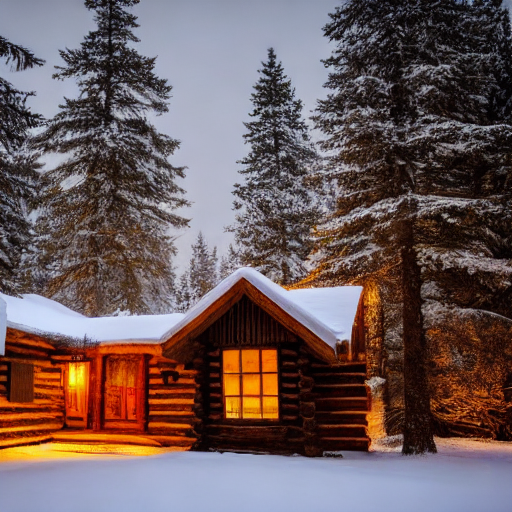} &
    \centering\includegraphics[width=\linewidth]{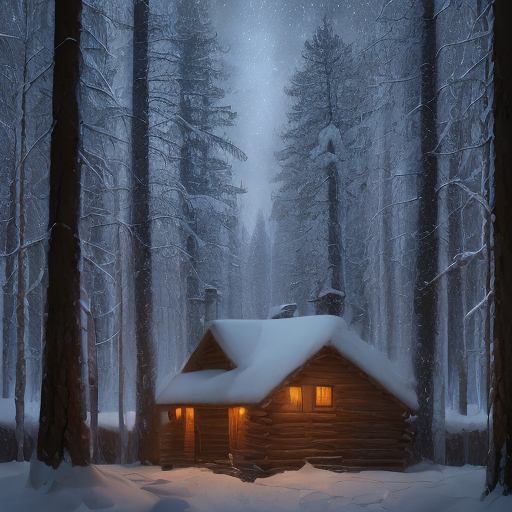} &
    \\
    & \centering\includegraphics[width=\linewidth]{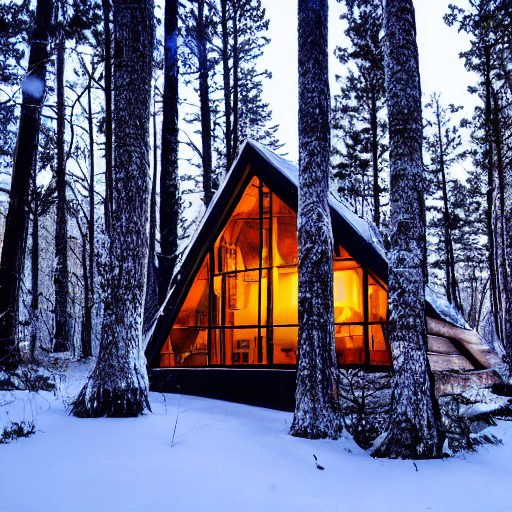} &
    \centering\includegraphics[width=\linewidth]{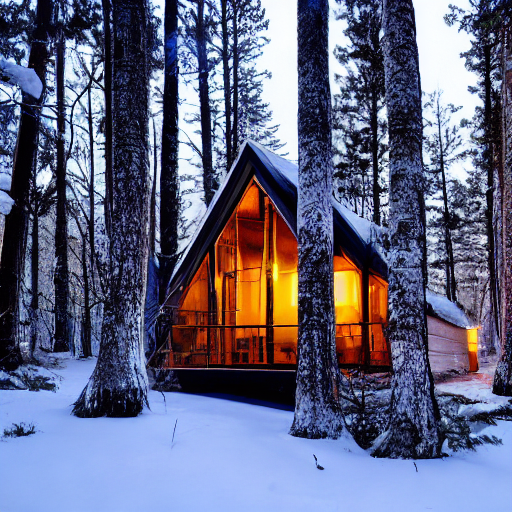} &
    \centering\includegraphics[width=\linewidth]{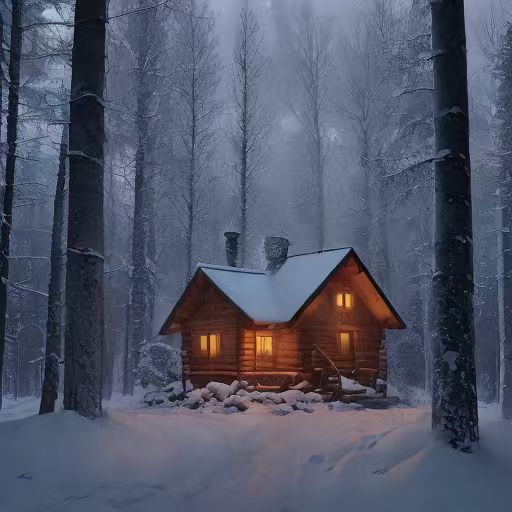} &
    \\

  \end{tabular}
  \vspace{-6pt}
  \caption{Comparison with LCM. We apply DRiffusion to LCM's teacher. \textit{A cozy cabin in a snowy forest at night, glowing windows.}}
  \label{fig:comp_distill}
\end{figure}

While parallelization and distillation methods each offer distinct benefits, we highlight the compelling advantages of our approach over distillation. First and foremost, DRiffusion is strictly training-free. Unlike distillation, which necessitates costly and model-specific training, our method serves as a plug-and-play algorithmic acceleration universally applicable to any pre-trained model. Furthermore, our approach excels in preserving both alignment and generation diversity. Distillation techniques, such as LCM \cite{luo2023latent}, often suffer from a noticeable distribution shift from the teacher model, leading to degraded diversity. In contrast, DRiffusion faithfully retains the original model's behavior. As illustrated in Figure \ref{fig:comp_distill}, while LCM deviates from the teacher's sampling trajectory and yields less diverse outputs, our method remains strictly aligned. Therefore, DRiffusion presents a more accessible acceleration paradigm, delivering substantial speedups without compromising the intrinsic quality or diversity of the original diffusion model.

\section{Future Works}

On the engineering side, an immediate direction is to extend DRiffusion to other generative modalities, such as video, policy generation, and audio. In these settings, diffusion models can serve as powerful and efficient generators of high-quality data for downstream tasks \cite{yao2025airroom}. 

On the algorithmic side, future work could further exploit the temporal flexibility of diffusion. Although DRiffusion is studied here in the multi-device setting, it may also accelerate single-device sampling by enabling larger batch sizes. Another important direction is to establish theoretical guarantees that bound the deviation of the draft trajectory from the original one, potentially enabling speculative-decoding-like methods \cite{leviathan2023fast} for diffusion sampling.

\end{document}